%% file: main.tex
\documentclass[a4paper,11pt]{article}

\usepackage{titling} 

\title{A Stochastic Process Model of Classical Search}
\author{
  \textbf{Dimitri Klimenko, Hanna Kurniawati, and Marcus Gallagher}\\
  The University of Queensland\\
  School of Information Technology and Electrical Engineering\\
  \texttt{dimitri.klimenko@uqconnect.edu.au}\\
  \texttt{\{hannakur,marcusg\}@uq.edu.au}
}

\usepackage{amsfonts} 
\usepackage{amsmath} 
\usepackage{amssymb} 
\usepackage{amsthm} 
\usepackage{oz} 
\usepackage[super]{nth}
\usepackage{mathrsfs}

\usepackage[T1]{fontenc} 
\usepackage{lmodern} 
\usepackage[UKenglish]{babel} 
\usepackage[babel=true]{microtype} 
\usepackage[autostyle]{csquotes} 

\usepackage{booktabs} 
\usepackage{multirow} 
\usepackage[table]{xcolor} 

\usepackage{float}
\usepackage{graphicx} 
  \graphicspath{{figs/}}
  \DeclareGraphicsExtensions{.pdf,.jpg,.png,.eps}

\usepackage{newclude} 

\usepackage{natbib}
\setcitestyle{authoryear}
\usepackage{hyperref}

\usepackage{algorithm} 
\usepackage[noend]{algpseudocode} 
\usepackage{wrapfig}

\usepackage[capitalise]{cleveref}
\crefname{line}{line}{lines}

\usepackage{xspace}
\input{macros}

\usepackage[title]{appendix}

\begin{document}
\maketitle

\input{allsections}

\clearpage


\bibliography{all}

\end{document}

%% file: macros.tex
\theoremstyle{plain}

\theoremstyle{definition}
\newtheorem{definition}{Definition}

\newcommand{\latinphrase}[1]{\textit{#1}}  
\newcommand{\etal}{\latinphrase{et~al.}\xspace}

\newcommand{\tTerm}{\textsc{Term}\xspace}
\newcommand{\tGoal}{\textsc{Goal}\xspace}

\newcommand{\illegalState}{\Gamma}

\newcommand{\allHists}{\mathcal{T}}
\newcommand{\finHists}{\mathcal{U}}
\newcommand{\illegalSet}{ \{ \illegalState \} }
\newcommand{\goalHists}{\allHists_{\tGoal}}

\newcommand{\featureSpace}{\mathcal{F}}
\newcommand{\featureSet}{F}

\newcommand{\featureSigmaAlgebra}{\Phi}
\newcommand{\fTermSet}{\featureSet_{\tTerm}}
\newcommand{\fGoalSet}{\featureSet_{\tGoal}}

\newcommand{\actionSet}{A}

\newcommand{\absFunc}{\phi}

\newcommand{\transFunc}{\kappa}

\newcommand{\problem}{\mathcal{G}}
\newcommand{\problemTuple}{
  \actionSet,
  \allHists,
  \goalHists,
  c
}

\newcommand{\absProblem}{\mathcal{\bar{G}}}
\newcommand{\absProblemTuple}{
  \actionSet,
  \featureSpace,
  \fGoalSet,
  x_0,
  \transFunc,
  \bar{c}
}

\newcommand{\finHistSpace}{A^{<\omega}}

\newcommand{\partCfg}{H, \absFunc_H}

\newcommand{\asmdpTuple}{
  \mathscr{S},
  \mathscr{A},
  \mathscr{T},
  \mathscr{R},
  s_0,
  \gamma
}
\newcommand{\asmdpState}{\partCfg}

\newcommand{\N}{\mathbb{N}}

\newcommand{\suchthat}{\mid}

\newcommand{\append}[2]{{#1}\cat{#2}}

\let\max\relax
\let\min\relax
\DeclareMathOperator*{\min}{min}
\DeclareMathOperator*{\max}{max}

\DeclareMathOperator*{\argmax}{arg\,max}

\newcommand{\expec}[1]{\mathbb{E} \left[ {#1} \right]}
\newcommand{\card}[1]{\left| {#1} \right|}


%% file: allsections.tex
\begin{abstract}
  Among classical search algorithms with the same heuristic information,
  with sufficient memory A* is essentially as fast as possible
  in finding a proven optimal solution.
  However, in many situations optimal solutions
  are simply infeasible,
  and thus search algorithms that trade solution quality for speed
  are desirable.
  In this paper, we formalize the process of classical search as a metalevel decision problem,
  the Abstract Search MDP.
  For any given optimization criterion, this establishes a well-defined notion of the best possible behaviour
  for a search algorithm and offers a theoretical approach to the design of algorithms for that criterion.
  We proceed to approximately solve a version of the Abstract Search MDP for anytime
  algorithms and thus derive a novel search algorithm,
  Search by Maximizing the Incremental Rate of Improvement (SMIRI).
  SMIRI is shown to outperform current state-of-the-art anytime search algorithms on a parametrized stochastic
  tree model for most of the tested parameter values.
\end{abstract}
\section{Introduction} \label{sec:intro}
Since the early 1960's, the general idea of \emph{best-first search} has been fundamental to the design of
a great many informed search algorithms.
Of particular note is A*~\citep{art:a_star},
which remains ubiquitous.
At a basic level, all best-first search algorithms work in the same way:
they incrementally build
a \emph{search tree},
in which nodes represent states in the problem's state space,
and edges represent state transitions. 
The key atomic task in this process is \emph{edge expansion},
which is the addition of a new node to the search tree via a previously unvisited edge. 

The design of any best-first search algorithm comes down to a single
fundamental question---how does the algorithm decide which edge to expand in each iteration?
In doing so there are trade-offs between minimizing cost of solutions and the amount of time
taken to find the solution.
Most work addresses the question
in an \emph{ad hoc} manner, 
on the basis of diverse heuristic arguments.
However, some researchers have taken more formal approaches to such questions;
of particular note is recent work~\citep{art:selecting} which formulates the Bayesian
selection problem as a \emph{metalevel decision problem}.
Unfortunately, the selection problem formalism is not sufficient to theoretically model the behaviour
of tree-searching algorithms.
In the words of Hay \etal,
``A more ambitious goal is extending the formalism to trees---in particular,
achieving better sampling at non-root nodes,
for which the purpose of sampling differs from that at the root.''

In this paper, we propose a decision-theoretic framework, the
\emph{Abstract Search Markov Decision Process} (ASMDP),
that achieves this ambitious goal for classical search problems
(in which results of actions are fully deterministic).
To show its benefit, we apply the framework to derive a novel model-based anytime search algorithm,
called Search by Maximizing the Incremental Rate of Improvement (SMIRI).
Experimental results on a random tree model
indicate that SMIRI offers state-of-the-art anytime performance in this domain.

\section{Related Work}\label{sec:related}
\subsection{Metareasoning and Models of Search}\label{ssec:metareasoning}
Although we would like search algorithms to exhibit perfect rationality, 
in the real world this is not not practical due to the high computational complexity of many
decision problems. 
To develop the most rational agent under limited computational resources, \emph{metareasoning} applies decision-theoretic principles.
The basic idea is simple---as with object-level decisions,
when choosing between computations an agent should select whichever one has the highest expected utility. 
Matheson~\citeyearpar{art:value_of_computation} showed that metareasoning can be formalized
by combining an object-level model with a model of the computations to form
the metalevel decision problem.
The Decision-Theoretic A* algorithm~\citep{art:dt_reasoning} applies metareasoning
to real-time problem-solving search,
but the utility estimates used in DTA*
cannot obey the standard axioms of probability and utility theory
and thus the approach lacks solid theoretical grounding~\citep{art:dt_reasoning}.
More recent work~\citep{art:selecting} has formalized the metalevel decision
problem for Bayesian selection problems. 
However, this formalism can only be applied
at the root node of a search tree,
and thus is not a full-fledged decision-theoretic framework for tree search algorithms.
In contrast, we propose a probabilistic approach that can be applied to tree search.


Probabilistic models of search have been proposed in the past,
but they have typically been used to analyze time complexity of pre-existing
algorithms~\citep{art:pearl_path,art:alpha_beta,art:nau_pathology}
rather than to formulate a metalevel decision problem.
One exception is the modelling by Mutchler~\citeyearpar{art:mutchler} of search with severely limited
edge expansions,
who showed theoretically that a simple best-first search algorithm was approximately optimal in a
simple random tree model.

The aforementioned prior works have offered interesting domains to study the behaviour of search algorithms,
and we continue this trend by formulating a novel random tree model that should be more representative of
classical search problems.
On the other hand another line of research focused on predicting the sizes of search trees has resulted in much more
realistic models of search problems,
primarily due to a focus on predicting real-world performance.
The earliest work on this problem is that of Knuth~\citeyearpar{art:predict_knuth};
later, Korf \etal~\citeyearpar{art:predict_korf} were able to accurately predict the number of nodes expanded
by the Iterative Deepening A* algorithm~\citep{art:ida_star} on a number of different problems.
This work was later extended by Zahavi \etal~\citeyearpar{art:cdp} to increase its accuracy,
by using conditional distributions over what Zahavi~\etal refer to as a \emph{type system}. 
In essence, the theoretical framework of this paper can be seen as building on the ``type system'' idea by viewing it
as a Bayesian model of the underlying search problem,
and building a metalevel decision problem (the Abstract Search MDP) on top of it.

\subsection{Existing Algorithms}\label{ssec:related_algorithms}
As mentioned in \cref{sec:intro}, different requirements with regard to solution quality and execution time
result in a spectrum of very different search algorithms to meet them.
The A* algorithm dominates one extreme within this spectrum---when used with an admissible heuristic,
A* ensures with complete certainty
that the solution it returns will be \emph{optimal}. 
Moreover, Dechter and Pearl \citeyearpar{art:optimality_a_star} demonstrate that 
A* is
(down to the tie-breaking criterion)
essentially the fastest general algorithm that can make this guarantee---other algorithms
can only do better by ``cheating'' on specific problem instances.
However, despite many adaptations of A*,
the issue of computational complexity remains hard to evade.
For many kinds of problems,
the time taken to find optimal solution will grow exponentially (or worse!) with the size of the problem,
no matter how good the search algorithm.

The only way to avoid this fundamental issue is to relax the optimality requirement,
e.g. to $w$-\emph{admissibility} which requires the solution to be within a factor $w$ of optimal. 
For a long time, the best-performing algorithm for this purpose was Weighted A*~\citep{art:wa_star};
this approach multiples the heuristic in A* by a constant factor of $w$,
resulting in an algorithm that typically runs faster than $A*$ but is $w$-admissible.
However, it tends to waste time exploring equally meritorious solutions in parallel,
even when any one of those solutions would be acceptable~\citep{art:kowalski_crit, art:a_eps_star}.
Explicit Estimation Search (EES)~\citep{art:ees} resolves this issue by maintaining multiple queues,
so as to maintain admissibility while focusing search effort on particular candidates.
An alternative suboptimality criterion is \emph{bounded-cost search},
which requires an algorithm to find a solution with cost strictly less than a fixed cost bound $C$.
The most prominent algorithm for this criterion is Potential Search~\citep{art:pts},
which aims to expand the node with the highest probability of having a path cost below the bound.

In this paper, we focus on the design of anytime algorithms~\citep{art:anytime},
which 
quickly find initial solutions and then gradually improve upon them.
A number of anytime heuristic search algorithms have been proposed in the literature;
the most basic is Anytime Weighted A*~\citep{art:anytime_a_star}, which is Weighted A* but continues to
search after a solution is found, and prunes nods that cannot lead to an improvement.
Anytime Repairing A*~\citep{art:ara_star} adapts this by also decreasing the weight every time a solution is found.
Finally, the latest state-of-the-art anytime algorithms have made further improvements by obviating the need for
tuning the weight parameters;
these are APTS/ANA*~\citep{art:pts,art:ana_star}, an anytime version of Potential Search,
and AEES~\citep{art:aees}, an anytime version of EES.
Experiments on random trees indicate that
SMIRI solidly outperforms these other algorithms.

\section{Framework} \label{sec:framework}

To define the ASMDP---the proposed metalevel decision problem for classical search---we first define the
\emph{search problem} and \emph{abstract search problem},
upon which it is built.


\subsection{The Classical Search Problem} \label{ssec:problem}
Here we use a simplified version of the 
\emph{extensive form game}~\cite[p. 89]{bk:gt_course},
so that the definition can easily be extended for all perfect information games,
including MDPs and stochastic games.
This definition is based on the idea of a \emph{history},
which is a sequence of actions.
In particular, for a given set of possible actions $A$,
we denote by $A^{\leq \omega}$ the set of all possible histories;
this can be split into the set of all infinite histories $A^\omega$
and the set of all finite histories $A^{< \omega}$.
Additionally, we use the symbol $\cat$ to denote appending an action to the end of a history,
e.g. $\append{h}{a}$.

At the core of any problem is a graph structure which we refer to as a \emph{search tree};
\begin{definition} \label{def:search_tree}
  A \emph{search tree} $\allHists$ is a set of histories which satisfies the following:
    \begin{itemize}
      \item The empty sequence or \emph{root} $\emptyseq$ is a member of
        $\allHists$.
      \item Any \emph{prefix} of a sequence in $\allHists$ is also in
        $\allHists$.
      \item Any infinite history all of whose prefixes
        are in $\allHists$ must also be in $\allHists$.
    \end{itemize}
\end{definition}
Thus a search tree represents a valid tree structure, in which the nodes of the tree are sequences
of actions and the edges are actions.

\begin{definition} \label{def:problem}
  A \emph{search problem}  is a tuple $\problem = (\problemTuple)$, where:
\begin{itemize}
  \item $\actionSet$ is the \emph{action space}---the finite set of all
    possible actions.
  \item $\allHists$ is a search tree, which we refer to as the \emph{complete search tree}
    of the search problem.
    Histories in $\allHists$ are called \emph{legal},
    and finite legal histories are called \emph{states},
    which are members of the \emph{state space}
    $\finHists = \allHists \cap \finHistSpace$.
  \item $\goalHists \subset \finHists$ is the set of \emph{solutions} to the search problem.
  \item $c : \finHists \to (0, \infty)$ is the \emph{step cost function},
    which maps each history to the cost of the last step taken.
    From $c$, the \emph{path cost function} $g$ is defined as 
    the sum of $c(n')$ over all histories $n'$ such that $n'$ is a prefix of $n$.
\end{itemize}
\end{definition}
Note that we represent the problem by a tree of histories, the \emph{complete search tree},
rather than the usual \emph{state-space graph}~\citep{bk:aima}.
Since the complete tree can be infinite, problems with cyclic state spaces can still be
correctly represented;
thus, this approach can still be used for searching on graphs, although it does limit
the ability of the theoretical model to account for redundant paths.

\subsection{The Abstract Search Problem} \label{ssec:abstract_problem}
Decision-theoretic analysis of classical search may appear to be straightforward,
but in reality there is a subtle issue that needs to be resolved~\citep{bk:right_thing}: 
any information obtained solely as the result of a computation is information that,
from the point of view of utility and probability theory,
that agent \emph{already had}. 
This fundamental issue is a rather difficult one;
it continues to be an area of active research,
sometimes referred to as the question of ``logical uncertainty''~\citep{art:logical_uncertainty}.

To address the issue, we reformulate the problem into a more standard question of
\emph{environmental uncertainty},
to which Bayesian probability and decision theory can be applied.
The idea is based on the simple observation that a search algorithm never actually
makes full use of the information available about the states in the search tree.
Therefore, a search algorithm only operates on an abstracted representation of a state which
we refer to as an \emph{abstract state}
residing in the feature space $\featureSpace$.
For generality, we consider the feature space to be an arbitrary measurable space, i.e.
$\featureSpace = (\featureSet, \featureSigmaAlgebra)$,
where $\featureSet$ is the set of all features, and $\featureSigmaAlgebra$ is a $\sigma$-algebra
of subsets of $\featureSet$,
which defines the measurable subsets of $\featureSet$.
The information available to the search algorithm is defined by the \emph{abstraction function}
$\absFunc : \finHists \to \featureSet$,
which maps a state to the features the search algorithm will actually use---a well-known example
is the Manhattan distance heuristic function in the sliding tile puzzle.
This idea can be seen as building upon the ``type system'' idea of
Lelis \etal~\citeyearpar{art:cdp2} by viewing
abstract states as residing in an arbitrary space,
and viewing the type system as representing a whole class of problems rather than just one.
This means that the same partial search tree can be consistent with distinctly different search problems
and thus the algorithm must be \emph{uncertain} about the true underlying search problem.
In this way, the process of search can be formally modeled as a process of \emph{Bayesian inference}
on a probability distribution over search problems. 
In particular, we view a search algorithm as starting with a \emph{prior} distribution over search problems,
which encodes initial knowledge about how features of states are related to one another;
this distribution is continually narrowed via Bayesian updating upon observing the features of states.

In order to make Bayesian inference amenable, we require the algorithm's prior knowledge to satisfy the
\emph{local directed Markov property} over the structure of the search tree.
In other words, the distribution of a subtree rooted at a particular node should be conditionally independent of
everything outside of that subtree, given the features of that node.
Although this might appear to be a severe restriction, it has intuitive appeal and still has the
capacity to model complex dependencies by adding extra features to the feature space.
Given the local directed Markov property, it follows that the prior can be defined entirely in terms of the
conditional distributions of child abstract states given parent abstract states,
i.e. a \emph{Markov kernel}. 
With this, we define the \emph{abstract problem} as:
\begin{definition} \label{def:abs_problem}
  An \emph{abstract problem} is a tuple $(\absProblemTuple)$, where
  \begin{itemize}
    \item $A$ is the action space, as per \cref{def:problem}.
    \item $\featureSpace = (\featureSet, \featureSigmaAlgebra)$
      is the \emph{feature space}.
      This is a measurable space, i.e. $\featureSet$ is the set of all possible abstract states,
      and $\featureSigmaAlgebra$ is a $\sigma$-algebra of subsets of $\featureSet$.
      This becomes the \emph{extended feature space}
      $\featureSpace^* = (\featureSet^*, \featureSigmaAlgebra^*)$
      by adding the \emph{illegal state} $\illegalState$
      to represent illegal actions.
    \item $\fGoalSet \subset \featureSet$ specifies the \emph{goal states}.
      These are analogous to set of solutions $\goalHists$ of a non-abstract
      search problem,
      but reside in the feature space rather than the state space.
    \item $x_0 \in \featureSet$ is the initial state;
      as with the goal states,
      this is analogous to the root history $\emptyseq$ but lies
      in the feature space.
    \item $\transFunc : ( \featureSet^* \times \actionSet ) \times \featureSigmaAlgebra^* \to [0, 1]$
      is the \emph{transition function}, a Markov kernel from
      state-action pairs to the extended feature space,
      such that:
      \begin{itemize}
        \item Any state/action pair must either map to the illegal state with probability 1,
          in which case it is \emph{illegal}, or with probability 0,
          in which case it is \emph{legal}.
        \item Any action leading out of the illegal state is also illegal,
          i.e. $\transFunc(\illegalState, \cdot, \illegalSet) \equiv 1$.
      \end{itemize}
      States with no legal actions are referred to as a \emph{sink states},
      which includes $\illegalState$ as well as the \emph{terminal states} $\fTermSet$.
    \item $\bar{c} : \featureSet \to (0, \infty)$ specifies \emph{step costs},
      as in \cref{def:problem}.
  \end{itemize}
\end{definition}

For a given abstract problem $(\absProblemTuple)$,
the induced \emph{tree generation process} 
is a stochastic process
$\{ \Psi_n \suchthat n \in \finHistSpace \}$,
where each $\Psi_n$ represents the abstract state associated with a history $n$.
\begin{definition}
  A \emph{realization} of the tree generation process (or, equivalently, the underlying abstract problem),
  is a pair $(\finHists, \phi)$,
  where $\finHists$ specifies which histories from $\finHistSpace$ are considered to be legal,
  and $\phi : \finHists \to \featureSet$ is an abstraction function (per the previous definition),
  which maps each legal history to an abstract state.
\end{definition}
Notably, a realization $(\finHists, \phi)$ fully defines a search problem per \cref{def:problem};
the action space is the same, and the complete search tree $\allHists$ is simply $\finHists$ with the addition of
any viable infinite histories.
The solutions to the realized search problem are
$\goalHists = \absFunc^{-1}(\fGoalSet)$,
and its step cost function is $c(n) = \bar{c}(\absFunc(n))$.
Thus, with the tree generation process we have fulfilled the task of constructing a well-defined prior
probability distribution over search problems.
Moreover, this distribution factorizes neatly into a tree of conditional distributions
(in essence, a finite or infinite Bayesian network);
since the features of a state are observed by the search algorithm,
a Bayesian update simply replaces unknown values for the random
variables $\Psi_n$ in the conditional distributions with known ones.

\subsection{The Abstract Search MDP (ASMDP)}
\label{ssec:asmdp}

In the context of the tree generation process as defined in the previous section, 
the process of search can be viewed as a Partially Observable Markov Decision Process (POMDP) in which the agent is unaware of the
true underlying problem, but can observe the abstract states within its search tree.

However, to avoid solving the full-blown POMDP directly, we instead apply the concept of
\emph{sufficient information states}~\citep{art:sufficient_stat} 
to define the ASMDP.
Due to the Markov assumption (\cref{ssec:abstract_problem}),
the structure of the search tree and the observed feature values within
that tree are a sufficient statistic.
\begin{definition}
  Given an action set $A$, a \emph{partial search tree} is a finite set of finite histories
  $H \subset \finHistSpace$
  that is also a search tree (per \cref{def:search_tree}).
  A \emph{partial realization} $(H, \absFunc_H)$ of an abstract problem $\absProblem$
  is a partial search tree $H$ and a mapping $\absFunc_H : H \to \featureSet$
  such that $x_0 = \absFunc_H(\emptyseq)$,
  and for any $\append{n}{a} \in H$ we have
  $\transFunc(\phi_H(n), a, \illegalSet) = 0$.
  In other words, it is a partial search tree labelled with abstract states;
  the condition on $\transFunc$ ensures that all histories in the partial realization
  must be legal,
  as illegal histories are not considered part of the search tree.
\end{definition}

Now, we can define the ASMDP.
\begin{definition} \label{def:asmdp}
  The \emph{ASMDP} for a given abstract problem $\absProblem$ is an
  MDP $(\asmdpTuple)$ over the space of partial realizations, where
  \begin{itemize}
    \item $\mathscr{S} = \{\partCfg \suchthat (\partCfg)
        \text{ is a partial realization of } \absProblem \}$ is the state space;
        the tree $H$ is the \emph{current search tree} of the searching agent.
    \item $\mathscr{A} = \finHistSpace$ is the action space.
      Legal actions in the ASMDP correspond to expanding previously unexpanded out-edges within the search tree $H$,
      i.e. an action $\append{n}{a}$ is legal if and only if $n \in H$, $\append{n}{a} \notin H$,
      and $\transFunc(\phi_H(n), a, \illegalSet) = 0$.
    \item $\mathscr{T}$ is the transition function, which is a Markov kernel.
      For any given state $s = (\asmdpState)$ and legal action $\append{n}{a}$, the next state $s'$ takes the form
      $s' = (H \cup \{\append{n}{a} \}, \phi_H \cup \{ (\append{n}{a}, X) \} )$,
      where $X$ is a random variable whose distribution is specified by the transition function of the abstract game,
      $\transFunc(\phi_H(n), a, \cdot)$.
    \item $\mathscr{R}$ is the reward function,
      which specifies the reward obtained by the searching agent for every time step.
      The choice of reward function will depend upon what we wish to optimize for in the design of the
      search algorithm.
    \item $s_0 = (\{ \emptyseq \}, \absFunc_0) $ is the \emph{initial state},
      where $\absFunc_0(\emptyseq) = x_0$.
      In other words, the initial state is a tree consisting of only the root node,
      labelled with the initial abstract state $x_0$.
    \item $\gamma$ is the \emph{discount factor},
      which depends on the type of search being modeled.
       \end{itemize}
\end{definition}
Within this formalism, an \emph{optimal search algorithm} corresponds directly to an \emph{optimal policy}
in the ASMDP.

For the case of anytime search, we view the search as being halted at an arbitrary time (which the agent is
uncertain about),
and evaluated based on the cost of the incumbent solution when the search is terminated.
In particular, we assume that there is a constant probability $p$ of the search terminating at time $k+1$
given that it has continued for $k$ steps.
Notably, rather than explicitly build this probability of halting into the ASMDP, we can model this by using
a discounting scheme where $\gamma = 1-p$,
and the reward is 
\begin{equation}
  \mathscr{R}(\asmdpState) = - \min \{ C_{max}, \min_{n \in \goalHists \cap H} g(n) \} ,
\end{equation}
i.e. the negation of the cost of the best solution present within the search tree,
up to an upper bound of $C_{max}$.
This reward is accumulated over time steps, but in conjunction with the discounting scheme the effect is that the
total expected reward for the MDP is essentially a weighted average of solution costs at different times,
weighted by how likely the search is to stop at that time.
\section{Approximately Solving the ASMDP}
\label{sec:solving}

Although the ASMDP results in a well-defined notion of what constitutes an \emph{optimal} search algorithm,
most general-purpose MDP solvers are too slow to handle the ASMDP due to state and action spaces that grow
hyper-exponentially with the number of steps taken.

%
However, by exploiting the local directed Markov property,
we can derive an efficient approximate solver.
The idea is analogous to \emph{index policies}~\citep{art:gittins} in multi-armed bandits:
it independently computes a single quantity for each action and selects the action with the greatest index.
Our index policy is derived based on the notion of \emph{incremental rate of improvement},
denoted as $r$, which is the rate at which the incumbent cost $C_{inc}$ improves over coming time steps
(thus ``rate of improvement''), but only in the near future (thus ``incremental'').
\footnote{Thayer \etal~\citeyearpar{art:aees} have argued against maximizing the rate of improvement,
but their argument applies to long-term improvement and not the short-term criterion used by SMIRI.}

To derive an approximate solution to the ASMDP, we first note that an optimal policy of the
ASMDP (i.e., an optimal search algorithm) 
should only ever expand edges that might either lead to a better solution, or be ``followed up'' on 
by the policy,
as these are the only ways to gain utility.
Consequently, the choice to expand an edge can be viewed as a ``macro-action'' with \emph{outcomes} $o \in O$;
which are either \emph{successes} $o \in O_s$, in which a better solution is found in the subtree for that edge
and the cost bound is reduced,
or \emph{failures} $o \in O_f$ in which the algorithm switches over to a different subtree without finding
an improved solution.
In particular, let $s$ be a state of the ASMDP with incumbent cost $C_{inc}$,
and $\pi$ be a policy that selects out-edge $\append{n}{a}$ in state $s$, and acts optimally otherwise.
We describe an \emph{outcome} $o \in O$ by three key parameters: its probability $p(o)$,
the subsequent resulting state $s'(o)$,
the amount $\Delta (o)$ by which the cost bound is reduced (if at all),
and the number of steps $t(o)$ taken to reach that outcome.
This means that any outcome $o$ is a series of $t(o)$ steps during which the agent receives a reward of $-C_{inc}$
on every step, until finally reaching a new ASMDP state $s'$ which might (or might not) have an improved
cost bound.
Then the value function of $\pi$ at state $s$ can be expressed as
\begin{equation}\label{eq:index_policy_value}
  V^{\pi}(s) = \sum_{o} p(o) \left[ \gamma^{t(o)} V^{\pi}(s'(o)) - C_{inc} \frac{1 - \gamma^{t(o)}}{1 - \gamma} \right].
\end{equation}
As there will typically be a large number of available out-edges with few leading to improvements,
for a failed outcome it is likely that
$V^{\pi}(s') \approx V^{\pi}(s)$.
On the other hand, if the solution is improved
the algorithm will gain by having a better solution for some time.
In principle, the amount of utility this gains could depend on the particular
cost bound and state, but as a rough approximation we assume that the utility gained
is proportional to the reduction in the cost bound, $\Delta$, i.e.
$V^{\pi}(s') \approx V^{\pi}(s) + \xi \Delta$ for some $\xi$.
Hence \cref{eq:index_policy_value} reduces to
\begin{equation}
  V^{\pi}(s) \approx \frac{
    \sum_{o} p(o) \left[ \gamma^{t(o)} \xi \Delta(o) - C_{inc} \frac{1 - \gamma^{t(o)}}{1 - \gamma} \right]
  }{
      1 - \sum_{o} p(o) \gamma^{t(o)}
    } .
\end{equation}
To further simplify this equation, we also assume that the times $t(o)$ are much smaller than the discounting horizon,
such that $\gamma^{t(o)} \approx 1 + t(o) \ln \gamma$.
In general, we expect that this assumption will be reasonable,
as the overall time spent searching should be much greater than the time spent focusing on
any one particular subtree for any one particular cost bound.
Thus, we expect that 
\cref{eq:index_policy_value} can be approximated as
\begin{equation}\label{eq:approx_value}
  V^{\pi}(s) \approx \frac{- \xi}{\ln \gamma} \frac{\sum_{o} p(o) \Delta(o)} {\sum_{o} p(o) t(o)} - \frac{C_{inc}}{1 - \gamma} .
\end{equation}
The incremental rate of imporvement $r$ is then the expected value of $\Delta$ divided by the expected value of $t$,
i.e.
\begin{equation} \label{eq:iri}
  r(s, \append{n}{a}) =  \frac{\sum_{o} p(o) \Delta(o)} {\sum_{o} p(o) t(o)} ,
\end{equation}
which is also the only term in \cref{eq:approx_value} that depends on $\pi$.

Consequently, a policy that always expands the edge with the maximal value of $r$ is an approximately optimal
policy for the ASMDP.
At this point, we have not yet precisely defined $r$, since the outcomes $o$ in \cref{eq:iri} depend on the
$r$-values themselves.
In particular, under our current definition a ``failure'' is when the algorithm switches over to a different
subtree without improving,
and this only happens when all unexpanded edges in the subtree being searched have lower $r$-values than
the best out-edge outside of that subtree.

As specified, there is no obvious way to calculate $r$ without knowing the optimal policy, since the stopping
criterion for failure introduces an interdependency between $r$-values of parallel out-edges.
However, we can derive an alternative form of $r$ without this flaw 
by specifying a different stopping criterion for failures $o \in O_f$.
In particular, define the \emph{thresholded incremental rate of improvement} $r_t$ as the incremental
rate of improvement obtained by searching optimally in a particular subtree until all out-edges
in that subtree are below the threshold $t$,
i.e. have $r_t(\cdots) < t$.
Finally, we define the \emph{peak incremental rate of improvement} $r^*$ as the maximum value of $r_t$
over all possible thresholds, i.e. $r^*(\cdots) = \max_{t \in [0, \infty) } r_t(\cdots)$.
This corresponds to expanding all edges in a subtree that are equal to or better than the root edge
of that subtree.
The thresholded and peak $r$-values still depend on other $r$-values, but only those of their children;
this results in a well-defined (via induction) notion of $r_t$ and $r^*$.

Moreover, it is clear that selecting whichever out-edge has the highest $r^*$ value will maximise the overall
incremental rate of improvement---a policy that selects another edge first is clearly dominated by a similar policy
that first expands the max-$r^*$ edge up to a threshold of $r^*$, and then does whatever the former policy did whenever
that fails.
This policy cannot be improved by including any edges below the threshold, or failing to include any above the threshold,
since the form of \cref{eq:iri} means that this necessarily results in a lower rate of improvement.

Finally, due to the local directed Markov property, the distribution of edges within the subtree for $\append{n}{a}$
depends only on $\phi(n)$ and $a$,
while $\Delta$ additionally depends on the relative cost bound $C_{inc} - g(n)$ within that subtree.
Since the peak rate of improvement does not depend on any other aspects of the ASMDP state $s$,
it can be expressed as a function $r^*(C_{inc} - g(n), \phi(n), a)$.
Consequently, a policy that always chooses to expand the out-edge $\append{n}{a}$ with the maximal value
of $r^*(C_{inc} - g(n), \phi(n), a)$ is an approximately optimal policy for the ASMDP.
This is essentially a best-first search on $r^*$, although it is important to note that the values of $r^*$ will
change whenever an improved solution is found, reducing the incumbent cost $C_{inc}$.
\section{Search by Maximizing the Incremental Rate of Improvement} 
\label{sec:algorithm}

\setlength{\textfloatsep}{5pt}
\begin{algorithm}[ht]
  \caption{SMIRI: computing $r^*$}\label{alg:smiri}
  \begin{algorithmic}[1]
    \For{C $\gets$ possible path costs from $0$ to $C_{max}$}
      \ForAll{legal $(x, a) \in \featureSpace \times \actionSet$}
        \State $e \gets (C, x, a)$
        \State $t_s(e) \gets 0; t_f(e) \gets 0$
        \State $edges_f(e, *) \gets 0$
        \If{$x \in \fGoalSet$}
          \State $p_s(e) \gets 1; \Delta (e) \gets C; r^*(e) \gets \infty$
          \State \textbf{continue}
        \EndIf
        \State $p_s(e) \gets 0; \Delta(e) \gets 0; r^*(e) \gets 0$
        \State $queue \gets \emptyset; freq(e, *, *) \gets 0$
        \ForAll {$y \in \featureSpace : p_y = \transFunc(x, a, \{ y \}) > 0 $}
          \State $p(y) \gets p_y; t(y) \gets 1$
          \ForAll{legal actions $b$ from $y$} 
            \State $e' \gets (C - \bar{c}(y), y, b)$
            \State $queue \gets queue \cup \{ (C-\bar{c}(y), y, b) \}$
            \State $freq(e, y, e') \gets 1$
          \EndFor
        \EndFor
        \While{$\max_{e' \in queue} r^*(e') \geq r^*(e)$}
          \State $e' \gets \argmax_{e' \in queue} r^*(e')$
          \State Remove $e'$ from $queue$
          \State \Call{ProcessDescendant}{e, e'}
          \State $t_f(e) \gets \sum_{y \in \featureSpace} p(y) t(y)$
          \State $r^*(e) \gets \Delta(e) / (t_s(e) + t_f(e))$
        \EndWhile
        \ForAll{$y, e' : m = freq(e, y, e') > 0$}
        \State $edges_f(e, e') \gets edges_f(e, e') + m \frac{p(y)}{1 - p_s(e)}$
        \EndFor
      \EndFor
    \EndFor

    \Procedure{ProcessDescendant}{$e, e'$}
      \ForAll{$y \in \featureSpace : m = freq(e, y, e') > 0$}
        \State $freq(e, y, e') \gets 0$
        \State $p_{suc} \gets 1 - (1 - p_s(e'))^m$ \label{line:p_suc}
        \State $t_{suc} \gets$ {
           $\sum_{k=0}^{m-1} \left[ t_s(e') + k t_f(e')\frac{p_s(e')}{1 - p_s(e')} \right] \left[ 1 - p_s(e')\right]^k$ \label{line:t_suc}
        }
        \State $p_s(e) \gets p_s(e) + p(y)p_{suc}$
        \State $t_s(e) \gets t_s(e) + p(y)\left(t_{suc} + p_{suc} t(y) \right)$
        \State $\Delta(e) \gets \Delta(e) + p(y) \Delta(e') p_{suc} / p_s(e') $
        \State $p(y) \gets p(y) (1 - p_{suc})$
        \State $t(y) \gets t(y) + t_f(e') / (1 - p_s(e'))$
        \ForAll{$e'' : m_2 = edges_f(e', e'') > 0$}
          \State Add $e''$ to the queue if it isn't there.
          \State $freq(e, y, e'') \gets freq(e, y, e'') + mm_2$
        \EndFor
      \EndFor
    \EndProcedure
  \end{algorithmic}
\end{algorithm}

In order to make practical use of the approximately optimal policy of \cref{sec:solving},
we need an efficient method for computing the peak incremental rate of improvement $r^*$
for each equivalence class of edges $e = (C, x, a)$;
where $C = C_{inc} - g(n)$ is the cost bound relative to the parent node of the out-edge,
$x$ is the abstract state of that node,
and $a$ is the action along the edge.
When path costs $g(\cdot)$ and abstract states $x$ are limited to finite sets,
\cref{alg:smiri} describes how to pre-compute $r^*$ in polynomial time.

The key idea is that since $g(n)$ can only increase along a path, the relative cost bound
$C$ will always decrease;
thus $r^*$ can be computed via dynamic programming along its first argument.
In particular, line 1 loops in increasing order of $C$, ensuring that $r^*$ values of descendant
edge types are always computed before their ancestors.

The inner loop on line 2 loops over all $x, a$ pairs for a given value of $C$, thus covering all
possible equivalence classes of edges $e = (C, x, a)$.
In order to compute the true $r^*$ value of $e$, it is necessary to find which descendants of $e$ fall
below the threshold, and sum the possible outcomes over all of those edges.
Lines 3-10 of the algorithm set all of the initial values for this calculation,
and then lines 11-16 initialise the possible descendants to include with the children of $(x, a)$,
using the transition function $\transFunc$ for the distribution of possible next-states $y$ resulting
from the $x, a$-transition.
In order to correctly find the optimal threshold value $r^*$, the value of $r^*(C, x, a)$
is calculated incrementally, gradually including all possible decendants $(C-\bar{c}(y), y, b)$ in order 
from the highest $r^*(\cdots)$ values to the lowest.
This is the core function of the loop over lines 17-22;
the process stops as soon as the current value of $r^*$ exceeds the $r^*$ values of all unused descendants.

For each equivalence class of edges $e = (C, x, a)$, the following variables are used to store cumulative values
that are updated as additional descendants are included in the threshold:
\begin{itemize}
  \item $p_s \approx \sum_{o \in O_s} p(o)$, the probability of success.
  \item $t_s \approx \sum_{o \in O_s} p(o) t(o)$, the unnormalized expected number of steps in a successful outcome.
  \item $t_f \approx \sum_{o \in O_f} p(o) t(o)$, the same for a failed outcome.
  \item $\Delta \approx \sum_{o \in O} p(o) \Delta(o)$: the \emph{expected improvement}.
  \item $r^* = \frac{\Delta}{t_s + t_f}$: the \emph{peak incremental rate of improvement}.
  \item $edges_f$: the expected frequencies of resulting edges given a failed outcome.
\end{itemize}
These quantities are updated by the \textsc{ProcessDescendant} subroutine, which adds the success and failure
statistics of one particular descendant $e'$ to the overall stats for $e$.
This also adds new descendants to the queue, since the descendants of $e'$ from $edges_f(e', *)$ are added.
Critically, any particular edge $e$ can only be evaluated once in lines 17-22, as they are evaluated in descending
order and the descendants of any edge always have lower $r^*$ values than the edge itself.

Also essential is that the quantities $p_s, t_s, \dots$ are are approximated.
Rather than computing them for hyper-exponentially many possibilities for 
expanded subtrees and taking expectations over those,
we view the subtrees as having only a single distinct outcome for each possible child abstract state $y$
resulting from the transition $(x, a)$,
i.e. all $y \in \featureSpace \suchthat \transFunc(x, a, \{ y \}) > 0$.
For this, we also store the quantities $p(y)$ and $t(y)$ to represent the total probability of failure and
aggregate time spent on failure within the subtree for $y$.
Within each of those subtrees, out-edges that may or may not appear are evaluated as though they occur the
expected number of times, represented
by non-integer exponents in \cref{line:p_suc,line:t_suc} of \cref{alg:smiri}.
Following this assumption, the expected frequencies for each type of edge $e$, descendant edge $e'$ and child state $y$ are stored in the
array $freq(e, y, e')$,
which is updated on lines 27 and 37.

Overall, if $D$ is the number of distinct possible values that $g(\cdot)$ can take,
then $W = D \card{\featureSpace} \card{\actionSet}$ is the number of possible distinct equivalence
classes of edges or $(C, x, a)$ tuples;
the final output of SMIRI is a lookup table consisting of at most $W$ entries.
SMIRI has worst-case space complexity of $O(W^2)$
(needed to store the descendant frequencies $edges_f$),
and worst-case time complexity of $O( \card{\featureSpace} W^3 \log W)$.
For homogeneous actions this reduces further as $W = D \card{\featureSpace}$,
and for typical cases with sparse transition functions the leading $\card{\featureSpace}$
term reduces to a constant $K$.
\section{Experiments} \label{sec:experiments}
\begin{table}[b]
  \centering
  \caption{Test parameters (harder cases first)}
  \label{tab:testcases}
  \begin{tabular}{ccrrcl}
    \toprule
    & \multicolumn{2}{c}{Model} & \multicolumn{3}{c}{Simulation} \\
    \cmidrule(lr){2-3} \cmidrule(l){4-6}
    & $p$ & $h_0$ & $C_{max}$ & $N$ & \multicolumn{1}{c}{$\gamma$} \\ 
    \midrule
    1 & 0.1 & 20 &      250 & $2\cdot10^6$ & $0.999~999$ \\ 
    2 & 0.2 & 100 &     300 & $2\cdot10^6$ & $0.999~999$ \\ 
    3 & 0.2 & 50 &      150 & $5\cdot10^5$ & $0.999~996$ \\ 
    4 & 0.2 & 20 &       80 & $1\cdot10^4$ & $0.999~8$   \\ 
    5 & 0.4 & 50 &       80 & $4\cdot10^3$ & $0.999~5$   \\ 
    6 & 0.6 & 50 &       70 & $1\cdot10^3$ & $0.998$     \\ 
    \bottomrule
  \end{tabular}
\end{table}

In order to evaluate the performance of SMIRI,
a random binary tree model $T(p, h_0)$ was designed to exhibit the typical characteristics of classical search problems.
We define $T$ as having a feature space of the natural numbers $\N$,
initial state $h_0$, goal state $0$, and legal actions $\textsc{Left}$ and $\textsc{Right}$ having
cost $\bar{c} \equiv 1$.
The transition function for $T(p, \cdot)$ from state $h$ with either action leads to state
$h-1$ with probability $p$ and $h+1$ with probability $1-p$.
This model has two key properties that make it a good testbed for anytime classical search.
First of all, the goal state can occur many times at many different depths in the tree,
resulting in solutions that vary widely in quality.
Secondly, for this problem $\phi$ is, in a natural manner,
the \emph{most accurate} admissible and consistent heuristic that can be constructed
from the feature space.
Thus, although SMIRI does not require any kind of admissible heuristic,
the model $T(p, h_0)$ offers a framework within which methods that rely on admissibility
can also be evaluated.
This model was implemented with 7 anytime classical search algorithms:
\begin{itemize}
  \item SMIRI, using $r^*$ as precomputed by \cref{alg:smiri}. 
  \item Anytime Potential Search (APTS/ANA*), as per Stern \etal~\citeyearpar{art:apts_ana_star},
    which selects the node with maximal $\frac{C - g(n)}{h(n)}$.
  \item Anytime Generalized Potential Search (AGPTS), as per Stern \etal~\citeyearpar{art:apts_ana_star}.
    For AGPTS, we explicitly precomputed a table for the \emph{potential} $PT_{C}(n)$ for all $n$ and $C$.
  \item Anytime Explicit Estimation Search (AEES), as per Thayer \etal~\citeyearpar{art:aees}.
    For AEES, we precomputed an unbiased inadmissible heuristic $\hat{h} \equiv \hat{d}$
    by calculating the expected value of the shortest-cost path from any node given $h$.
  \item Anytime Repairing A* (ARA*) per Likachev \etal~\citeyearpar{art:ara_star};
    with weights 5, 3, 2, 1.5, 1 per Richter \etal~\citeyearpar{art:rwa_star}.
\end{itemize}
Performance was evaluated in terms of solution cost vs number of edges expanded in the search,
thus the details of hardware and algorithm implementations should not be relevant to the results.
Although using step counts instead of time neglects the relative overhead of the different algorithms,
the average time per step was very similar for most of them;
the only notable exception was EES, which can be several times slower per expansion if node generation
operations are cheap~\citep{art:ees}.

\begin{figure}[t!]
  \centering
  \label{fig:profiles}
  \includegraphics{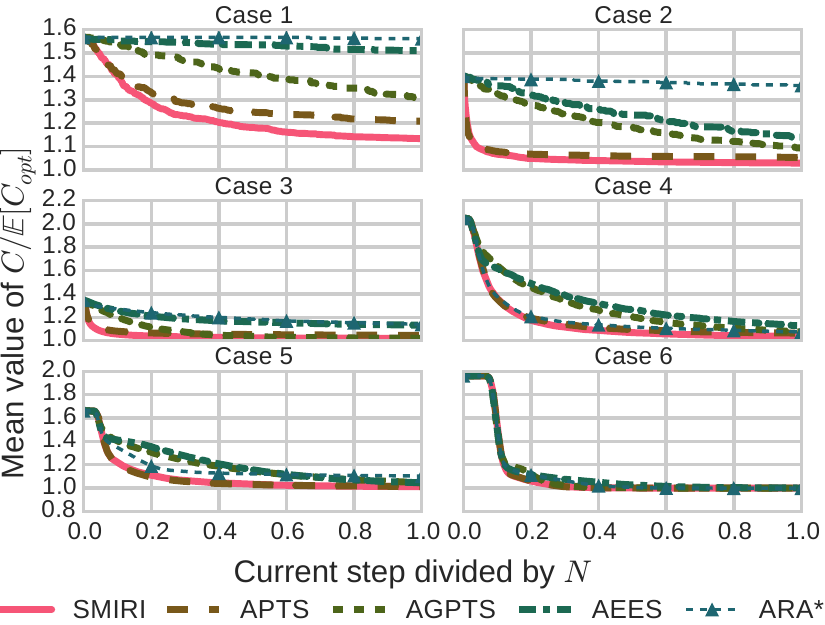}
  \caption{Mean solution quality profiles of the tested anytime algorithms.}
\end{figure}

Each of the algorithms was run on 6 different sets of parameter values,
which are specified in detail in \cref{tab:testcases}.
The cost bound $C_{max}$ was chosen to make it relatively easy to find an improved initial solution
in each test case,
whereas the edge expansion limit $N$ was chosen to allow the higher-quality algorithms sufficient time to
closely approach optimal solutions where possible.
The discount factor was then chosen as $\gamma = 1 - 2 / N$ ($\gamma^N \approx 0.135$).
Tables for $r^*$, $PT_{C}$ and $\hat{h}$ were precomputed,
as was the expected cost of an optimal solution $\expec{C_{opt}}$,
which was used to estimate suboptimality as calculating optimal solution costs for particular instances
quickly becomes impractical.
For each test case 1000 instances 
(except in test cases 1 and 2, which only used 100)
of $T(p, h_0)$ were created as \emph{generative models}---this is necessary since instances of $T(p, h_0)$
are typically \emph{infinite} and cannot be explicitly instantiated.
Each algorithm was then run on each instance of the random tree model for $N$ iterations,
recording the sequences of improved solutions and the time step at which each was attained.
Note that each model instance was shared between all of the algorithms, which significantly reduces variance in the
relative performance of different algorithms due to luck on particular model instances.

\Cref{fig:profiles} depicts the profile of incumbent solution cost $C$ vs.
the number of edges expanded for each algorithm and test case.
For easier comparison, the $y$-axes are presented as the estimated suboptimality, i.e. $C / \expec{C_{opt}}$,
while the $x$-axes are the number of edge expansions divided by the maximum $N$ allowed for each test case.
Finally, \cref{tab:costs} depicts mean \emph{discounted total cost};
these values are normalized, with the total discounted cost for each policy being divided by the
expected cost of a hypothetical ``perfectly rational'' anytime algorithm which has a solution of mean quality
$\expec{C_{opt}}$ after $0$ time steps.
\begin{table}[t]
  \centering
  \caption{Normalized discounted total cost. Lower values are better; lowest per testcase bolded}
  \label{tab:costs}
  \begin{tabular}{ccccccc}
    \toprule
    & \multicolumn{6}{c}{Test case number} \\
    \cmidrule(l){2-7}
    Algo. & 1 & 2 & 3 & 4 & 5 & 6 \\
    \midrule
        SMIRI  &\bf{ 1.28} &\bf{ 1.06} &\bf{ 1.05} &\bf{ 1.25} &    1.16   &    1.20    \\
        APTS   &    1.33   &    1.08   &    1.07   &    1.27   &\bf{ 1.15} &\bf{ 1.20}  \\
        AGPTS  &    1.46   &    1.25   &    1.11   &    1.42   &    1.28   &    1.21    \\
        AEES   &    1.54   &    1.29   &    1.21   &    1.45   &    1.29   &    1.22    \\
        ARA*   &    1.57   &    1.38   &    1.22   &    1.28   &    1.24   &    1.21    \\
    \bottomrule
  \end{tabular}
\end{table}
\section{Discussion} \label{sec:discussion}
The decreasing differences between algorithms indicate that the test cases become progressively easier from 1 to 6,
with a large decrease in difficulty between $p=0.6$ and $p=0.4$;
this suggests the model may have a \emph{complexity transition} at $p=0.5$,
as with the random tree model of Karp and Pearl~\citep{art:pearl_path}
Overall, SMIRI and APTS/ANA* are quite clearly the best-performing algorithms,
particularly for the hardest test cases.
By contrast, ARA* performed poorly for these cases,
suggesting that for difficult problems it is much
better to use algorithms that don't rely on parameter tuning.

As can be seen in cases 1, 2, and 3, APTS appears to level off at higher-cost solutions than SMIRI,
and thus is likely could be far slower to attain solutions of similar quality than SMIRI.
More critically, there is a significant disparity between APTS and AGPTS, which is a significant theoretical
concern for APTS,
because if one follows the theoretical derivation given by Stern~\etal~\citeyearpar{art:apts_ana_star}
AGPTS does the ``correct thing'' in always selecting for maximum potential $PT_{C}$,
whereas the linear-relative assumption APTS relies on clearly fails for the random tree model we have used.
In general the results for APTS in this domain are much better than those of AGPTS,
although for later timesteps in case 3 AGPTS overtakes APTS in solution quality.
The most likely explanation for this issue is one that has been raised by Stern~\etal themselves:
although AGPTS correctly selects for maximum potential,
this is not actually the correct thing for an anytime search algorithm to do.
In particular, considering only the probability of a solution in a given subtree neglects the
expected \emph{search effort} to find a solution there.
For example, a node with 10 20-step step branches, each of which has a 50\% chance of being a solution path,
has a 99.9\% being part of a solution,
whereas a node with a single 5-step branch that has a 90\% chance of being a solution path has a 90\% chance
of being part of a solution.
Clearly the potential of the first node is higher, but a reasonable search algorithm should always expand the
latter node first as it is likely to lead to a better solution than the former,
and more quickly to boot.

The idea of minimising search effort is one of the key motivations for AEES~\citep{art:ees},
but unfortunately it performed significantly worse than SMIRI or APTS and sometimes ARA*.
It is difficult to say why this occurs, but it is likely that the problem is related to the nature of the
inadmissible heuristic $\hat{h}$---it appears that, despite being an unbiased estimator,
the statistical properties of $\hat{h}$ and/or the underlying random tree model cause issues for AEES.
Nevertheless, AEES has one key advantage, which is the use of a \emph{distance-to-go} estimate $d(n)$ to better
estimate search effort;
this does not give any advantage in our test domains since every edge has unit cost, and thus
$\hat{d} \equiv \hat{h}$.
However, SMIRI also makes an explicit distinction between estimates of cost and distance,
as evidenced by the distinct quantities $\Delta$, $t_s$ and $t_f$ in \cref{alg:smiri}.

In summary, SMIRI was the best overall algorithm in this test domain, with a strong edge over
all other algorithms except APTS (over which it has only a slight edge).
The very narrow losses to APTS in test cases 5 and 6 suggest that both algorithms may be acting quite close to optimally;
the approximations made in \cref{sec:solving,sec:algorithm} are likely to be the cause of suboptimality in SMIRI.
More critically, the results for AGPTS demonstrate that the good performance for APTS comes \emph{in spite} of the
justification by Stern~\etal for selecting on the basis of potential, and not \emph{because} of it.
This is highlighted by the benchmarking results of Thayer~\etal~\citeyearpar{art:aees},
which show AEES significantly outperforming APTS in all but one domain.
On the whole, SMIRI has shown solid results in the synthetic benchmark, while also lacking some of the theoretical
shortcomings of APTS.
\section{Conclusion} \label{sec:conclusion}
Overall, the experimental results for SMIRI are quite promising,
although SMIRI is not without limitations---it requires pre-processing time,
and is limited to domains in which the possible path costs (up to $C_{max}$)
and the abstract states are finite sets.
However, such pre-processing costs can become insignificant when compared to a large search problem
or amortized over many searches that share an abstract model.
Furthermore, one possible approach that could resolve both of these issues is functional approximation methods,
which could be used to capture the structure of $r^*$.
A greater problem with the results is that they are limited to an artificial problem domain,
rather than real-world problems or typical benchmark problems from the literature.
It would be very informative to examine the performance of SMIRI in more realistic problem domains,
and particularly ones with non-uniform edge costs to check the effectiveness of SMIRI's use of estimated
search effort.

However, the main purpose of this paper is not to evaluate the SMIRI algorithm, but rather to demonstrate
the effectiveness of applying explicit metareasoning techniques to the problem of classical search.
The benchmarks, although limited, are solid evidence that metareasoning in general, and the ASMDP in particular,
are a useful tool for developing better search algorithms.
In particular, this paper demonstrates how formal decision-theoretic methods can be used to formulate the problem
of heuristic search as a metalevel decision problem with a well-defined optimal solution,
rather than relying on ad hoc heuristic arguments to justify a particular method.
Furthermore, although the benchmarking here is limited, many of the latest results in probabilistic models of
deterministic search~\citep{art:scp} indicate that such models \emph{can} be effectively applied in
estimating the performance of search algorithms,
and there is little reason to believe that the additional step of applying metareasoning to such models is
fundamentally flawed.

Although SMIRI is derived from a particular application of the ASMDP framework to anytime search,
the ASMDP framework applies more generally to other kinds of design criteria for search algorithms.
Moreover, the core framework should not be difficult to extend to search with non-classical elements such as
adversaries or stochastic environments.
A far more serious limitation is the assumption of a tree structure and the local directed Markov property in the
construction of the tree generation process,
as this restricts the ASMDP's ability to handle correlations that are induced by cycles in
graph-structured problems.
Nevertheless, the success of probabilistic prediction methods for tree search in domains with cyclic
graphs~\citep{art:scp} indicates that such models can be useful even if they don't account for those factors.
Moreover, despite this limitation of the theory, SMIRI can still be adapted to function as a graph search algorithm
in the usual way,
by adding a closed list and re-expanding edges when necessary.
On the whole, both the theoretical and experimental results of this paper are quite promising, and indicate a clear
need both for further experimental results with the SMIRI algorithm,
as well as a broader theoretical investigation of the ASMDP framework.